\documentclass[letterpaper, 10 pt, conference]{ieeeconf}  

\IEEEoverridecommandlockouts                              

\usepackage{graphics} 
\usepackage{epsfig} 
\usepackage{mathptmx} 
\usepackage{times} 
\usepackage{amsmath} 
\usepackage{amssymb}  

\usepackage{booktabs}
\usepackage{multirow} 

\usepackage{hyperref}
\DeclareGraphicsExtensions{.pdf, .png}
\title{\LARGE \bf
Legged Robot State Estimation Using Invariant Neural-Augmented Kalman Filter with a Neural Compensator
}

\author{Seokju Lee, Hyun-Bin Kim, and Kyung-Soo Kim
\thanks{This work was supported by Agency for Defense Development (ADD) - Grant funded by Defense Acquisition Program Administration (DAPA) in 2020 (UD230004TD). \textit{(Corresponding author: Kyung-Soo Kim)}}
\thanks{The authors are with the Mechatronics, Systems and Control Lab (MSC Lab), Department of Mechanical Engineering, Korea Advanced Institute of Science and Technology (KAIST), Yuseong-gu, Daejeon 34141, Republic of Korea (e-mail: \{dltjrwn0322, youfree22, kyungsookim\}@kaist.ac.kr)}}%

\begin{document}

\maketitle
\thispagestyle{empty}
\pagestyle{empty}

\begin{abstract}


This paper presents an algorithm to improve state estimation for legged robots. Among existing model-based state estimation methods for legged robots, the contact-aided invariant extended Kalman filter defines the state on a Lie group to preserve invariance, thereby significantly accelerating convergence. It achieves more accurate state estimation by leveraging contact information as measurements for the update step. However, when the model exhibits strong nonlinearity, the estimation accuracy decreases. Such nonlinearities can cause initial errors to accumulate and lead to large drifts over time. To address this issue, we propose compensating for errors by augmenting the Kalman filter with an artificial neural network serving as a nonlinear function approximator. Furthermore, we design this neural network to respect the Lie group structure to ensure invariance, resulting in our proposed Invariant Neural-Augmented Kalman Filter (InNKF). The proposed algorithm offers improved state estimation performance by combining the strengths of model-based and learning-based approaches. Project webpage: \url{https://seokju-lee.github.io/innkf_webpage}
\end{abstract}

\section{INTRODUCTION}
Significant research has focused on legged robots capable of traversing challenging terrains such as stairs, slopes, and slippery surfaces \cite{grandia2023perceptive, lee2020learning, miki2022learning}. These robots are also used for terrain mapping, where SLAM algorithms rely heavily on accurate state estimation. Errors in state estimation can cause localization drift, leading to mapping failure \cite{wisth2019robust}. This issue is especially critical when exteroceptive sensors (e.g., LiDAR, cameras) are unavailable, requiring state estimation using only proprioceptive sensors like IMUs and encoders.

State estimation methods have consisted of model-based and learning-based approaches. Among model-based methods, the state-of-the-art Invariant Extended Kalman Filter (InEKF) defines the state on a Lie group to improve convergence speed. It estimates states by incorporating contact foot kinematics as measurements \cite{hartley2020contact}. However, InEKF assumes static contact, leading to significant errors when a slip occurs. To address this, methods like slip rejection \cite{kim2021legged} and Invariant Smoothers (IS) \cite{yoon2023invariant} have been proposed.

InEKF updates states using a first-order approximation for computing the matrix $\mathbf{A}_t$, which fails to account for nonlinearities in error dynamics, leading to estimation drift:
\begin{equation} \frac{\text{d}}{\text{dt}} \boldsymbol{\xi}_t = \mathbf{A}_t \boldsymbol{\xi}_t + \text{Ad}_{\bar{\mathbf{X}}_t} \mathbf{w}_t. \end{equation} \begin{equation} \boldsymbol{\eta}_t^r = \exp(\boldsymbol{\xi}_t) \approx \mathbf{I}_d + \boldsymbol{\xi}_t^{\wedge}. \end{equation}

Meanwhile, learning-based approaches like KalmanNet \cite{revach2022kalmannet}, Neural Measurement Networks (NMN) \cite{youm2024legged}, and methods that concurrent train state estimators and controllers \cite{ji2022concurrent} have been introduced. However, these methods still suffer from estimation errors, highlighting the need for further legged robot state estimation improvements.

In this paper, we first adopt the InEKF and propose the Invariant Neural-Augmented Kalman Filter (InNKF), which augments the InEKF with a neural network to compensate for errors in the InEKF-estimated state. Since errors in the states estimated by the InEKF arise because the error dynamics exhibit significant nonlinearity due to various factors (such as slip, model uncertainty, and sensor noise), a compensation step is added using a neural network as a nonlinear function approximator. A neural compensator capable of producing elements of the SE$_{2}$(3) Group is introduced to ensure that the neural network also preserves invariance. This neural compensator outputs the error  
$\hat{\textbf{E}_{t}}^{r}=\bar{\textbf{X}}_{t}^{+}\bar{\textbf{X}}_{t}^{-1} \in \text{SE}_\text{2}(3)$
as its result. Ground truth data is obtained from simulations, and the error between the InEKF estimates and the ground truth is calculated. This error is then used as a data label for training. At every time-step $t$, this value compensates the updated estimate  
$\bar{\textbf{X}}_{t}^{+}$, resulting in  
$\bar{\textbf{X}}_{t}^{++}=\hat{\textbf{E}_{t}}^{r^{-1}}\bar{\textbf{X}}_{t}^{+} \in \text{SE}_2(3)$. Our main contributions are as follows:




\begin{itemize}
\item We propose a state estimator that combines the strengths of model-based and learning-based approaches, introducing an additional compensation step via a neural compensator to compensate errors caused by nonlinearities in the model-based estimated state.
\item During the state estimation process, all states were defined on a Lie group to enhance the convergence speed of the model. To achieve this, the neural network was designed to output elements of the Lie group.
\item Training data was collected from scenarios in the simulation where state estimation errors were likely to occur. The proposed method exhibited high state estimation performance on terrains where nonlinearities caused significant errors.
\end{itemize}

This paper is organized as follows. Section \ref{section:related_work} introduces general approaches for contact estimators, assuming that no foot-mounted sensors are used, and provides an overview of legged robot state estimators. Section \ref{section:method} explains the proposed method in detail, while Section \ref{section:experiments} presents the experimental setup and results. Finally, Section \ref{section:conclusion} concludes the paper and discusses future work.

\section{RELATED WORK}
\label{section:related_work}

\subsection{Contact Estimator}


Legged robots have contact between their feet and the ground during locomotion. This contact can introduce errors in state estimation, which can be addressed by utilizing contact information as measurements during the update step of the InEKF \cite{hartley2020contact}, as shown in (\ref{eq:meas}).

\begin{equation}
    \mathbf{Y}_t^\top = \begin{bmatrix} h_p^\top (\tilde{\boldsymbol{q}}_t) & 0 & 1 & -1 \end{bmatrix}
    \label{eq:meas}
\end{equation}
where $h_p(\tilde{\boldsymbol{q}}_t)$ represents the forward kinematics of contact position measurements.

Estimating the contact state is a critical issue in legged robot state estimation. While the actual contact state can be measured using force/torque or pressure sensors, many legged robots do not utilize such sensors. Consequently, it is necessary to estimate the contact state using only IMUs and encoders. Various contact estimators for legged robots have been studied, including methods based on Convolutional Neural Networks (CNN) \cite{pmlr-v164-lin22b} and probabilistic estimation using Hidden Markov Models (HMM) \cite{hwangbo2016probabilistic}.

Among these approaches, Marco Camurri \cite{Camurri2017ContactEstimation} proposed a method where, for each contact point $i$, the contact force $\mathbf{f}_i$ can be estimated based on robot dynamics and joint torques, as given in (\ref{eq:est_force}). The normal force component, perpendicular to the contact surface, is then computed via the inner product, as expressed in (\ref{eq:normal_force}). Using this value, the contact probability $P_i$ is determined through logistic regression, as shown in (\ref{eq:contact_prob}).

To use this information as measurements for InEKF, it is necessary to compute the covariance. The covariance is derived based on the variation in the estimated normal force from the previous time step, as formulated in (\ref{eq:contact_cov}). Finally, based on the computed contact probability, if the probability exceeds a predefined threshold $\theta$, the contact state is set to true, as defined in (\ref{eq:contact_state}).

In this study, we use this contact estimator to design a legged robot state estimator.

\begin{equation}
\mathbf{f}_i = - \mathbf{J}_i^\top (\mathbf{J}_i \mathbf{J}_i^\top)^{-1} (\boldsymbol{\tau}_i - \mathbf{g}_i)
\label{eq:est_force}
\end{equation}
where $\mathbf{J}_i$ is the contact Jacobian matrix at contact point $i$, defined at the joint positions, $\boldsymbol{\tau}_i$ is the joint torque applied at contact point $i$, and $\mathbf{g}_i$ is the force computed at contact point $i$ using inverse dynamics.

\begin{equation}
f_{i,\text{normal}} = \mathbf{f}_i \cdot \mathbf{n}_i
\label{eq:normal_force}
\end{equation}
where $\mathbf{n}_i$ represents the normal vector of the contact surface at contact point $i$.

\begin{equation}
P_i = \frac{1}{1 + \exp{(-\beta_1[i] f_{i,\text{normal}} - \beta_0[i])}}
\label{eq:contact_prob}
\end{equation}
where $\beta_0[i], \beta_1[i]$ are the regression coefficients.

\begin{equation}
\Sigma_i = k \cdot (f_{i,\text{normal}} - f_{i,\text{normal, prev}})^2
\label{eq:contact_cov}
\end{equation}

\begin{equation}
\text{S}_i = 
\begin{cases} 
\text{true}, & \text{if } P_i \geq \theta \\
\text{false}, & \text{otherwise}
\end{cases}
\label{eq:contact_state}
\end{equation}
where $S_i$ represents the contact state at contact point $i$.

\subsection{Legged Robot State Estimator}


Legged robots can perform state estimation using either model-based or learning-based approaches. Model-based methods typically rely on Kalman filters, and due to the nonlinear dynamics of legged robot systems, the extended Kalman filter (EKF) is commonly employed. By leveraging leg kinematics and IMU data, EKF variants have been developed to address specific challenges. For instance, the quaternion-based EKF (QEKF) \cite{bloesch2013state} resolves singularity issues and improves numerical stability using a quaternion-based mathematical model. To enhance the convergence speed of QEKF, the Invariant EKF (InEKF) \cite{hartley2020contact} defines states on a Lie group to ensure invariance. This study adopts the contact-aided InEKF, which is widely used in legged robot state estimators.

For learning-based approaches, neural networks have been incorporated into the Kalman filter framework to compute covariances, as in KalmanNet \cite{revach2022kalmannet}. Other methods include state estimators that learn contact events across various terrains for integration into InEKF \cite{lin2021legged}, approaches that simultaneously train legged robot control policies and state estimators \cite{ji2022concurrent}, and Pronto \cite{camurri2020pronto}, which utilizes learned displacement measurements \cite{buchanan2022learning}. Additionally, the Neural Measurement Network (NMN) \cite{youm2024legged} employs neural networks to estimate measurement values for use in InEKF.

This study also employs neural networks; however, it proposes a state estimator by augmenting a model-based approach with neural networks.

\section{METHOD}
\label{section:method}

In this study, the Neural Compensator (NC) is augmented to the InEKF, enabling an additional compensation of the error in the model-based updated estimated state to obtain the final estimated state $\bar{\textbf{X}}_{t}^{++}$. During this process, the NC outputs an element of the SE$_2$(3) Group, ensuring that invariance is consistently maintained in the estimated states.

\subsection{SE$_2$(3) Group Generation Network}

The Neural Compensator is a neural network augmented to the InEKF, designed to consider invariance by outputting elements on the Lie group. To achieve this, the focus was placed on values defined in the Lie algebra, i.e., the tangent space, when designing the neural network. First, the state $\textbf{X}_t$ is defined in (\ref{eq:state}) as an element of $\mathbb{R}^{5 \times 5}$. If the generators of the SE$_2$(3) group's Lie algebra ($\mathfrak{se}_2$(3)) are expressed as $\textbf{G}_i (i \in \{1, 2, \cdots, 9\})$, then the elements of $\mathfrak{se}_2$(3) can be represented as the linear combination of the generators, as shown in (\ref{eq:linear_comb}).

\begin{equation}
\textbf{X}_t = \begin{bmatrix}
        \textbf{R}_t & v_t & p_t \\
        \textbf{0}_{1,3} & 1 & 0 \\
        \textbf{0}_{1,3} & 0 & 1 \\
    \end{bmatrix} \in \text{SE}_2(3)
    \label{eq:state}
\end{equation}

\begin{align}
    \begin{aligned}
        \xi &= (\omega, v, p)^\top \in \mathbb{R}^9 \\
        \sum_{i=1}^{9} \xi_i G_i &\in \mathfrak{se}(3), \quad \text{where } 
        \xi_i = 
        \begin{cases} 
        \omega_i & \text{if } i \leq 3, \\
        v_{i-3} & \text{if } 4 \leq i \leq 6, \\
        p_{i-6} & \text{if } i > 6.
        \end{cases}
    \end{aligned}
    \label{eq:linear_comb}
\end{align}


\begin{figure}
    \centering
    \includegraphics[width=\linewidth]{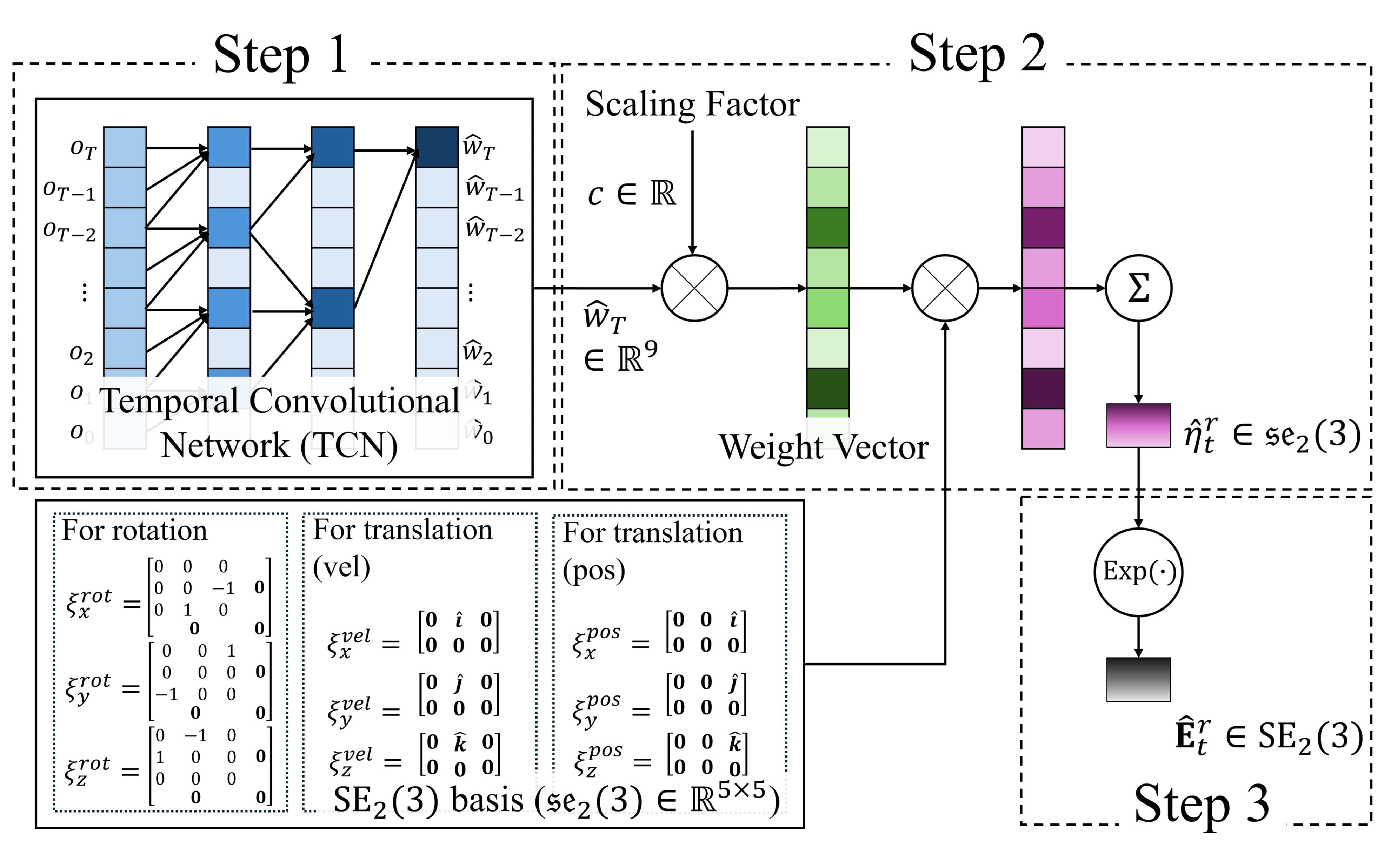}
    \caption{The SE$_2$(3) Group Generation Network (SEGGN) consists of three steps: (1) generating the weights for the elements of $\mathfrak{se}_2$(3), (2) performing a linear combination of the TCN outputs with the elements of $\mathfrak{se}_2$(3), and (3) applying exponential mapping to the results from step (2).}
    \label{fig:se3_gn}
\end{figure}
\begin{figure*}[htb!]
    \centering
    \includegraphics[width=0.95\linewidth]{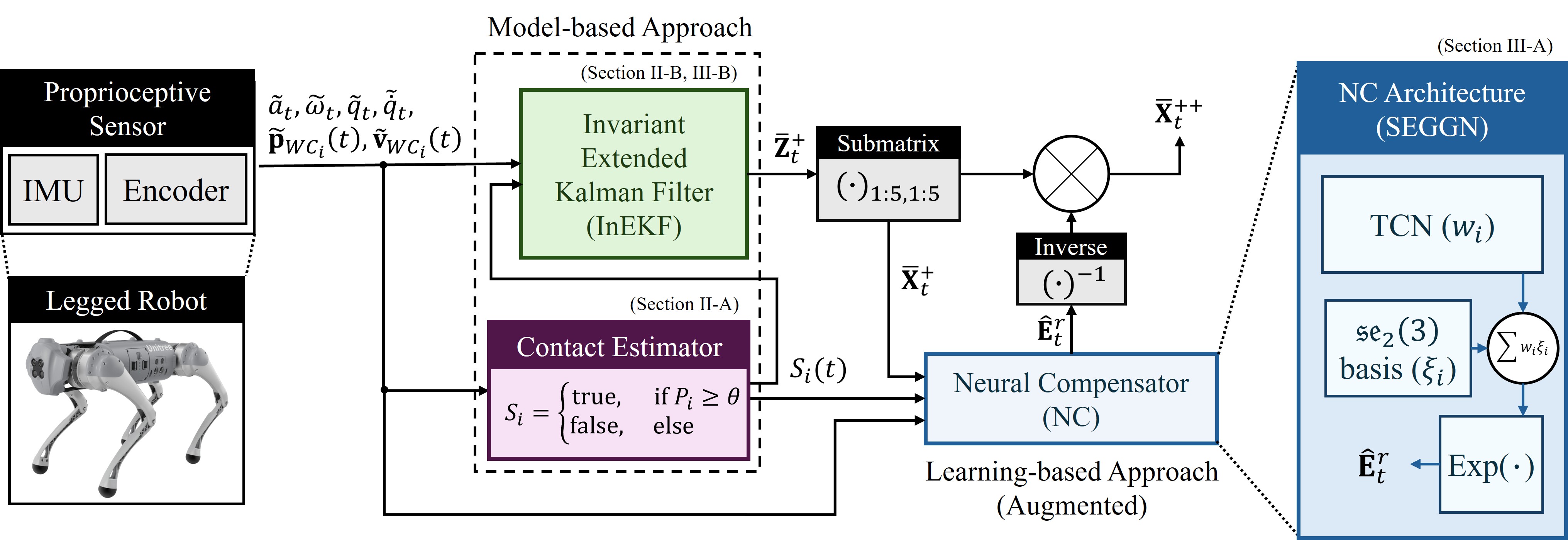}
    \caption{The overall architecture of the Invariant Neural-Augmented Kalman Filter (InNKF). InNKF involves the predict and update step of InEKF and Neural Compensator (NC) that reduces error values obtained which is designed to SE$_2$(3) Group Generation Network (SEGGN).}
    \label{fig:innkf_archi}
\end{figure*}

Using the linear combination of the generators, the neural compensator, as shown in Fig. \ref{fig:se3_gn}, consists of three main steps. First, a temporal convolutional network (TCN) \cite{bai2018empirical} takes as input a time-series sequence that includes linear acceleration, angular velocity, joint position, joint velocity, foot position, foot velocity, contact state, and the estimated state from the InEKF over the previous 49 time-steps as well as the current time step, and it produces 9 outputs. Second, these outputs are combined linearly with the generators of $\mathfrak{se}_2$(3) to obtain an element of $\mathfrak{se}_2$(3), denoted as $\hat{\eta}_{t}^{r}$. Finally, by applying the exponential mapping described in (\ref{eq:exponential_map}), the result is mapped to an element of the SE$_2$(3) group, $\hat{\textbf{E}}_t^r$, which represents the estimated error. This value acts as a compensation term for the estimated state in the InEKF.
\begin{align}
    \begin{aligned}
        \text{exp}(\cdot) &: \hat{\eta}_{t}^{r^{\wedge}} \mapsto \hat{\textbf{E}}_t^r \in \text{SE}_2(3) \\
        \text{Exp}(\cdot) &: \hat{\eta}_{t}^{r} \mapsto \hat{\textbf{E}}_t^r \in \text{SE}_2(3)
    \end{aligned}
    \label{eq:exponential_map}
\end{align}

The hidden layers of the TCN structure were configured as [128, 128, 128, 256, 256], with a kernel size of 2, a dropout rate of 0.5, and the ReLU activation function. The supervised learning process of the TCN was conducted as follows. The labeled dataset, which consists of ground truth error, was collected from simulation data. Specifically, ground truth values and estimated values from the InEKF were saved during the simulation. The target value for computing the loss was defined as $\textbf{E}_{t}^{r} = \bar{\textbf{X}}_{t}^{+} \textbf{X}_{t}^{-1}$, and training was performed using this target value. Based on the tangent space values obtained through the neural network, exponential mapping was applied to map them to the SE$_2$(3) group. The loss function was calculated accordingly since the target value was also defined on the SE$_2$(3) group.

The loss function was divided into a rotation part and a translation part. The rotation part corresponds to the top 3x3 matrix of the SE$_2$(3) matrix, and the rotation loss was computed using the Frobenius norm and the geodesic distance. The translation part, which consists of velocity and position (i.e., the vectors in the 4th and 5th columns of the SE$_2$(3) matrix), was used to compute the translation loss via Mean Squared Error. The total loss was calculated by weighting the rotation and translation losses, and training was performed. The Adam optimizer was used for training, with a learning rate 5e-4. 

The InEKF operated at a frequency of 500 Hz, and inputs with the same frequency were provided for training. The TCN window time was set to 0.1 seconds, corresponding to a window size of 50, allowing the model to receive sequences of 50 time-steps as input. The window shift size was set to 1, resulting in a shift of 0.002 seconds for training.
\subsection{Invariant Neural-Augmented Kalman Filter}

\begin{figure}
    \centering
    \includegraphics[width=0.9\linewidth]{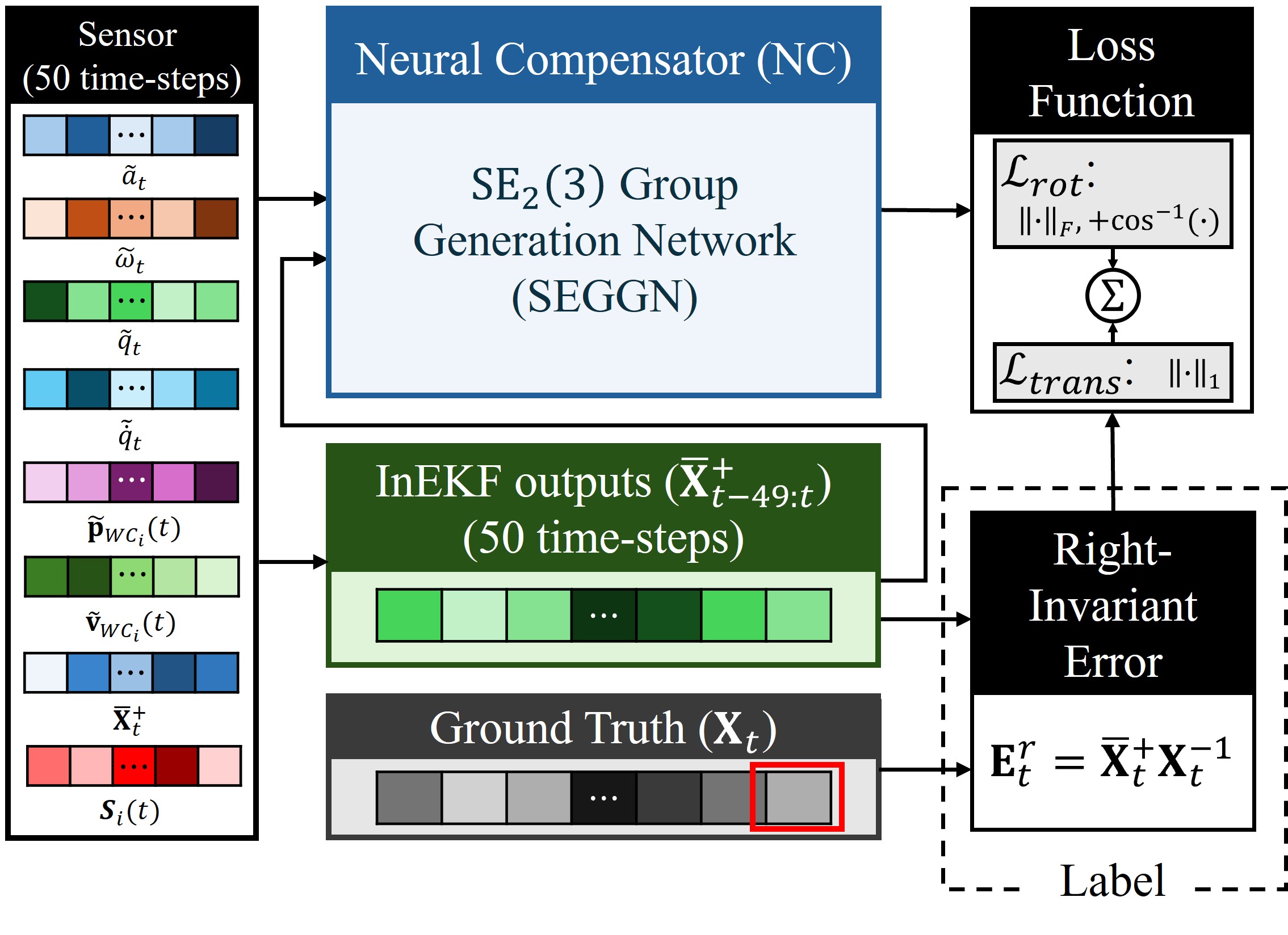}
    \caption{Training process of the Neural Compensator (NC): The dataset is collected in 50 time-step sequences, where state estimates are obtained at each time step using the InEKF. The right-invariant error is computed and labeled only for the final time step. Based on this labeled error, the SEGGN is then trained using the output and a loss function.}
    \label{fig:training_process}
\end{figure}
The Neural Compensator introduced in Section III-A is augmented into the InEKF \cite{hartley2020contact} to form the Invariant Neural-Augmented Kalman Filter (InNKF). The structure of InNKF is shown in Fig. \ref{fig:innkf_archi}, where the state estimation is first performed using a model-based approach. In this process, two types of states are defined. The state of the InEKF is expressed as $\textbf{Z}_t \in \text{SE}_{N+2}(3)$, as shown in (\ref{eq:Z_t}), while the state of the InNKF is expressed as $\textbf{X}_t \in \text{SE}_2(3)$. Here, $\textbf{X}_t$ is a $5 \times 5$ submatrix in the top-left corner of $\textbf{Z}_t$.

\begin{equation}
\mathbf{Z}_t \triangleq
\begin{bmatrix}
\mathbf{X}_t & {}^{\text{w}}\mathbf{p}_{\text{WC}_1}(t) & {}^{\text{w}}\mathbf{p}_{\text{WC}_2}(t) & \cdots & {}^{\text{w}}\mathbf{p}_{\text{WC}_N}(t)\\
\mathbf{0}_{1,5} & 1 & 0 & \cdots & 0 \\
\mathbf{0}_{1,5} & 0 & 1 & \cdots & 0 \\
\vdots  & \vdots  &  \vdots & \ddots & \vdots\\
\mathbf{0}_{1,5} & 0 & 0 & \cdots & 1\\
\end{bmatrix}
\label{eq:Z_t}
\end{equation}
where $\textbf{X}_t$ represents the (\ref{eq:state}), and ${}^{\text{w}}\mathbf{p}_{\text{WC}_i}(t)$ represents the each contact point for measurements of InEKF.

The IMU measurements used in the InEKF can be modeled using additive white Gaussian noise (AWGN), as expressed in (\ref{eq:IMU}).
\begin{equation}
    \begin{aligned}
    \tilde{\boldsymbol{\omega}}_t &= \boldsymbol{\omega}_t + \mathbf{w}_t^{\text{g}}, 
    \quad \mathbf{w}_t^{\text{g}} \sim \mathcal{N}\big(\mathbf{0}_{3 \times 1}, \Sigma^{\text{g}} \delta(t - t')\big) \\
    \tilde{\mathbf{a}}_t &= \mathbf{a}_t + \mathbf{w}_t^{\text{a}}, 
    \quad \mathbf{w}_t^{\text{a}} \sim \mathcal{N}\big(\mathbf{0}_{3 \times 1}, \Sigma^{\text{a}} \delta(t - t')\big)
    \end{aligned}
\label{eq:IMU}
\end{equation}
where $\mathcal{N}$ is a Gaussian distribution and $\delta(t - t')$ represents the Dirac delta function. Additionally, in the contact measurements, the velocity of the foot can also be modeled by considering potential slip and additive white Gaussian noise (AWGN). The measured foot velocity can thus be expressed as shown in (\ref{eq:foot_vel}).
\begin{equation}
{}^{\text{w}}\tilde{\textbf{v}}_C(t) = {}^{\text{w}}\textbf{v}_C(t) + \mathbf{w}_t^{\text{v}}, 
\quad \mathbf{w}_t^{\text{v}} \sim \mathcal{N}\big(\mathbf{0}_{3 \times 1}, \Sigma^{\text{g}} \delta(t - t')\big)
\label{eq:foot_vel}
\end{equation}

The system dynamics derived from the IMU and contact measurements are expressed as shown in (\ref{eq:system_dyn}):
\begin{equation}
\begin{aligned}
    &\frac{\text{d}}{\text{dt}} \textbf{R}_t = \textbf{R}_t(\tilde{\boldsymbol{\omega}}_t - \mathbf{w}_t^{\text{g}}) \\
    &\frac{\text{d}}{\text{dt}} \textbf{v}_t = \textbf{R}_t(\tilde{\boldsymbol{a}}_t - \mathbf{w}_t^{\text{a}}) + \textbf{g} \\
    &\frac{\text{d}}{\text{dt}} \textbf{p}_t = \textbf{v}_t \\
    &\frac{\text{d}}{\text{dt}} {}^{\text{w}}\textbf{v}_C(t) = \textbf{R}_th_R(\tilde{q}_t)(-\mathbf{w}_t^{\text{v}})
\end{aligned}
\label{eq:system_dyn}
\end{equation}
where $h_R(\tilde{q}_t)$ denotes the orientation measurements of the contact frame in IMU frame by forward kinematics using encoder measurements, $\tilde{q}_t$. This equation can be represented in matrix form as shown in (\ref{eq:predict}), which allows the computation of the predict step.
\begin{equation}
    \begin{aligned}
        \frac{\text{d}}{\text{dt}}\bar{\textbf{Z}}_t &= f_{u_t}(\bar{\textbf{Z}}_t) \\
        \frac{\text{d}}{\text{dt}}\textbf{P}_t &= \textbf{A}_t\textbf{P}_t + \textbf{P}_t\textbf{A}_t^\top + \bar{\textbf{Q}}_t
    \end{aligned}
    \label{eq:predict}
\end{equation}
where $f_{u_t}(\cdot)$ represents the matrix form of (\ref{eq:system_dyn}), $\textbf{P}_t$ is the covariance matrix, and $\textbf{A}_t$ and $\bar{\textbf{Q}}_t$ are obtained from the linearization of the invariant error dynamics as shown in (\ref{eq:lin_err_dyn}):
\begin{equation}
    \begin{aligned}
        \mathbf{A}_t &= 
\begin{bmatrix}
    \mathbf{0} & \mathbf{0} & \mathbf{0} & \mathbf{0} \\
    \mathbf{g}_\times & \mathbf{0} & \mathbf{0} & \mathbf{0} \\
    \mathbf{0} & \mathbf{I} & \mathbf{0} & \mathbf{0} \\
    \mathbf{0} & \mathbf{0} & \mathbf{0} & \mathbf{0}
\end{bmatrix} \\
\overline{\mathbf{Q}}_t &= \text{Ad}_{\overline{\mathbf{Z}}_t} \text{Cov}(\mathbf{w}_t) \text{Ad}_{\overline{\mathbf{Z}}_t}^\top.
    \end{aligned}
    \label{eq:lin_err_dyn}
\end{equation}
with $\mathbf{w}_t=\text{vec}(\mathbf{w}_t^\text{g}, \mathbf{w}_t^\text{a}, \textbf{0}_{3,1}, h_R(\tilde{q}_t)\mathbf{w}_t^{\text{v}})$.

The predicted state, $\bar{\textbf{Z}}_t$, is updated using the measurements, $\textbf{Y}_t$ from (\ref{eq:meas}), along with the covariance, as shown in (\ref{eq:update}).
\begin{equation}
    \begin{aligned}
        \bar{\textbf{Z}}_t^+ &= \text{exp}(\textbf{K}_t\Pi_t\bar{\textbf{Z}}_t\textbf{Y}_t)\bar{\textbf{Z}}_t \\
        \textbf{P}_t^+&=(\textbf{I}-\textbf{K}_t\textbf{H}_t)\textbf{P}_t(\textbf{I}-\textbf{K}_t\textbf{H}_t)^\top+\textbf{K}_t\bar{\textbf{N}}_t\textbf{K}_t^\top
    \end{aligned}
    \label{eq:update}
\end{equation}
where the $\Pi_t$ defines the auxiliary selection matrix, which is the $[\textbf{I} \quad \textbf{0}_{3, 3}]$, and the gain $\textbf{K}_t$ is obtained from (\ref{eq:kalman_gain}):
\begin{equation}
 \textbf{K}_t=\textbf{P}_t\textbf{H}_t^\top(\textbf{H}_t\textbf{P}_t\textbf{H}_t^\top+\bar{\textbf{N}}_t)^{-1},
    \label{eq:kalman_gain}
\end{equation}
$\textbf{H}_t$ is defined to $[\textbf{0}_{3, 3} \quad \textbf{0}_{3, 3} \quad -\textbf{I} \quad \textbf{I}]$, and $\bar{\textbf{N}}_t$ is computed by $\bar{\textbf{R}}_t\textbf{J}_p(\tilde{q}_t)\text{Cov}(\textbf{w}_t^{q})\textbf{J}_p^\top(\tilde{q}_t)\bar{\textbf{R}}_t^\top$.


Now, the state $\bar{\textbf{Z}}_t^+$, estimated through the InEKF, is used to extract a $5 \times 5$ submatrix from the top-left corner, which represents only the robot base's state. The remaining values correspond to the contact positions of the robot's foot and are unnecessary for compensation; therefore, the submatrix is extracted. This extracted submatrix becomes $\bar{\textbf{X}}_t^+$, which undergoes additional compensation to obtain the final estimated state.


Using the SE$_2$(3) Group Generation Network (SEGGN) designed in Section III-A, a Neural Compensator (NC) is added. The NC is trained on a dataset obtained from a simulation in which a single trajectory traverses four different terrains (stairs, slope, random uniform, discrete obstacle) for 100 seconds. The dataset included the ground truth state $\textbf{X}_t$, sensor data such as linear acceleration ($\tilde{a}_t$), angular velocity ($\tilde{\omega}_t$), joint position and velocity ($\tilde{q}t, \tilde{\dot{q}}t$), foot position and velocity ($\tilde{\textbf{p}}_{WC_i}(t), \tilde{\textbf{v}}_{WC_i}(t)$), contact state ($\text{S}_t$), and the estimated state $\bar{\textbf{X}}_t^+$ from the InEKF. This dataset was used to train the SEGGN through the supervised learning process shown in Fig. \ref{fig:training_process}.

During the training process, the ground truth error, defined as $\textbf{E}_t^r = \bar{\textbf{X}}_t^+ \bar{\textbf{X}}_t^{-1} \in \text{SE}_2(3)$, is used as the label. The loss is calculated between this value and the output of the SEGGN.

The training process employed two loss functions to separately handle rotation and translation. For the rotation part, the loss function in (\ref{eq:rotation_loss}) was formulated using the geodesic distance and Frobenius norm. The Frobenius norm captures element-wise differences between the ground truth and network output, while the geodesic distance ensures proper evaluation of rotational discrepancies on \text{SO}(3). This combination makes (\ref{eq:rotation_loss}) effective for training. The Frobenius norm is defined in (\ref{eq:frob}).
\begin{equation}
\mathcal{L}_{\text{fro}} = \frac{1}{2} \| \textbf{R}_1 - \textbf{R}_2 \|_F^2,
\label{eq:frob}
\end{equation}
where \( \| \textbf{R}_1 - \textbf{R}_2 \|_F = \sqrt{\sum_{i,j} \left( \textbf{R}_1[i,j] - \textbf{R}_2[i,j] \right)^2} \) denotes the element-wise difference between the matrices. This metric ensures computational efficiency while providing a measure of proximity in Euclidean space.

On the other hand, the geodesic distance represents the minimal angular displacement required to align \(\textbf{R}_1\) with \(\textbf{R}_2\) on the rotation group \(\text{SO}(3)\). It is computed as (\ref{eq:geodesic}):
\begin{equation}
\mathcal{L}_{\text{geo}} = \arccos \left( \frac{\mathrm{trace}\left( \textbf{R}_1^T \textbf{R}_2 \right) - 1}{2} \right),
\label{eq:geodesic}
\end{equation}
where \( \mathrm{trace}(\cdot) \) is the sum of the diagonal elements of the matrix. This measure captures the intrinsic rotational difference and is particularly meaningful in applications sensitive to angular deviations.

\begin{equation}
\mathcal{L}_{\text{rot}}(\textbf{R}_1, \textbf{R}_2) = \alpha \mathcal{L}_{\text{fro}} + \beta \mathcal{L}_{\text{geo}}
\label{eq:rotation_loss}
\end{equation}
where \( \alpha \) and \( \beta \) are weighting factors that balance the contributions of the Frobenius norm and the geodesic distance. This formulation ensures both computational simplicity and rotational accuracy.

The loss function for the translation part is formulated using the L1 loss, as shown in (\ref{eq:translation_loss}).
\begin{equation}
\mathcal{L}_{\text{trans}} = \| \textbf{v}_1 - \textbf{v}_2 \|_1 + \| \textbf{p}_1 - \textbf{p}_2 \|_1,
\label{eq:translation_loss}
\end{equation}

The final loss function is constructed using (\ref{eq:rotation_loss}) and (\ref{eq:translation_loss}), as expressed in (\ref{eq:loss_fn}), and is used for training.
\begin{equation}
\mathcal{L} = c_1 \mathcal{L}_{\text{rot}} + c_2 \mathcal{L}_{\text{trans}}
\label{eq:loss_fn}
\end{equation}
where $c_1$ and $c_2$ denote the coefficients of the loss functions.

The SEGGN trained using (\ref{eq:loss_fn}) is employed as the NC and augmented into the InEKF. The label used for training the SEGGN is given by $\textbf{E}_t^r = \bar{\textbf{X}}_t^+ \bar{\textbf{X}}_t^{-1}$, and the output of the SEGGN can therefore be expressed as $\hat{\textbf{E}}_t^r$. Since the SEGGN's output represents the error that must be corrected in the state estimated by the InEKF, it can be formulated as shown in (\ref{eq:compensate}).
\begin{equation}
    \bar{\textbf{X}}_t^{++}=\hat{\textbf{E}}_t^{r^{-1}}\bar{\textbf{X}}_t^+
    \label{eq:compensate}
\end{equation}

where $\bar{\textbf{X}}_t^{++}$ represents the compensated state derived from the estimated state by the InEKF. This compensated state serves as the final state estimation of the InNKF, with the InEKF and NC operating in parallel. The InEKF continuously computes estimates based on proprioceptive sensor data, while the NC uses the estimates from the InEKF and proprioceptive sensor values as inputs to predict the estimated error. The NC operates with a delay of one time step compared to the InEKF and corrects the InEKF's estimated states using its estimated error. The compensated state is not reused in the InEKF but is solely used as the final state estimate.

\section{EXPERIMENTS}
\label{section:experiments}
This study aims to enhance the state estimation performance of legged robots by evaluating rotation, position, and velocity estimation. Simulations provide ground truth data for direct performance assessment, but testing on a single simulator may lead to overfitting. To ensure generalization, the estimator was further evaluated in a different dynamic simulator. To introduce significant nonlinearities, trajectories were designed on terrains with sudden rotational changes. Additionally, real-world performance was tested in an uneven terrain environment.
\subsection{Performance Analysis of SE$_2$(3) Group Generation}
\begin{figure}[t!]
    \centering
    \includegraphics[width=0.95\linewidth]{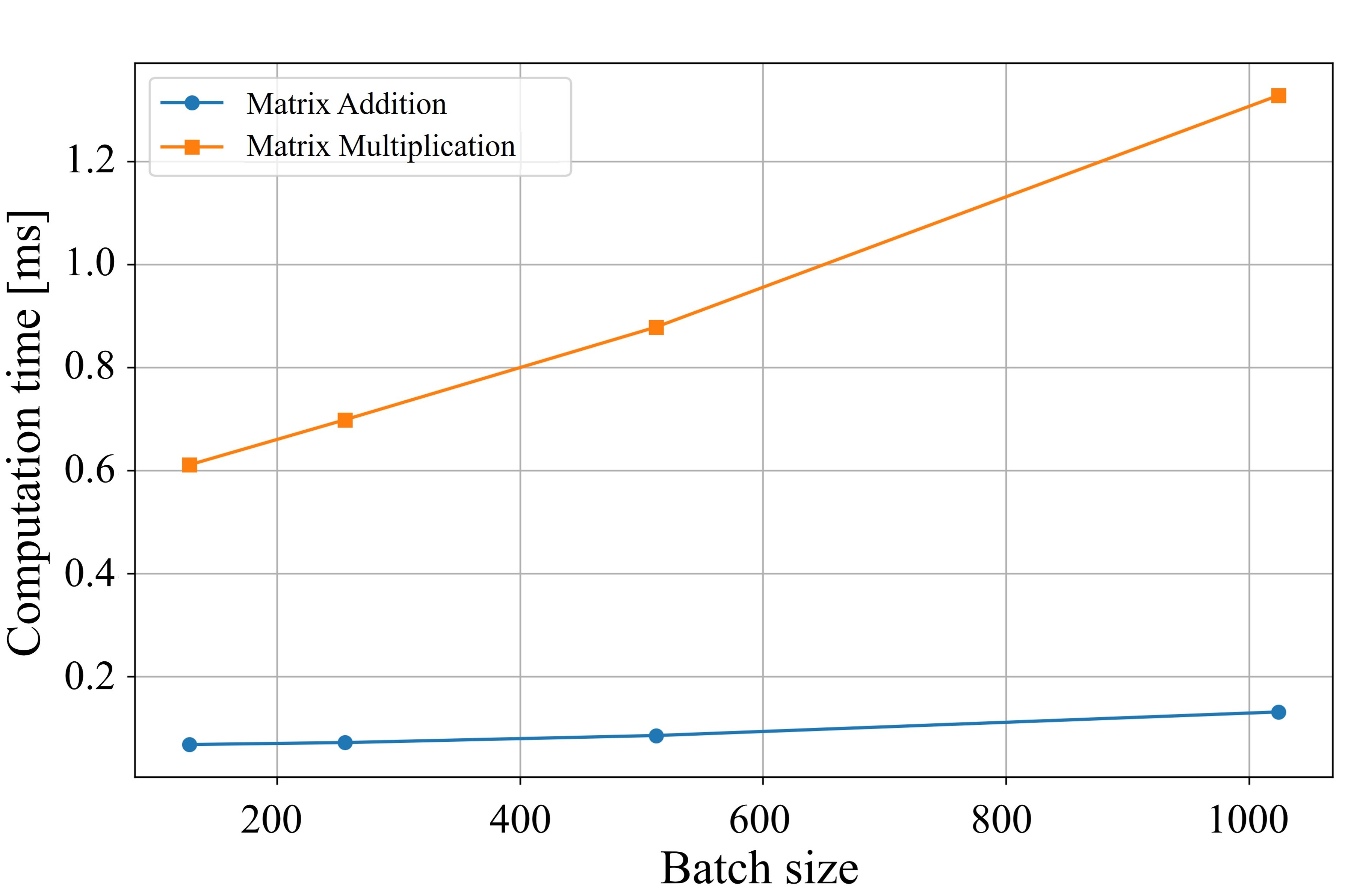}
    \caption{Comparison of addition and multiplication operation time depending on the batch size. This graph represents the efficiency of computation in lie algebra.}
    \label{fig:computation_time}
\end{figure}

The neural network used in this study does not perform operations directly on the Lie group but operates in the Lie algebra, followed by exponential mapping. Since the Lie group is defined on a smooth manifold, performing group operations requires matrix multiplication. In contrast, the Lie algebra, defined in the tangent space, only requires matrix addition for operations. Consequently, obtaining the output in the Lie algebra and applying exponential mapping significantly improves training efficiency. Table \ref{table:add_mul_computation} and Fig. \ref{fig:computation_time} show the computation time according to batch size. While matrix addition has a time complexity of O\((n^2)\), matrix multiplication has a time complexity of O\((n^3)\). This highlights the advantage of obtaining outputs in the tangent space, where only matrix addition is required, to enhance training efficiency. A difference of approximately 0.1 times in computation time was observed, confirming significantly faster performance.

\begin{figure}[t]
    \centering
    \includegraphics[width=\linewidth]{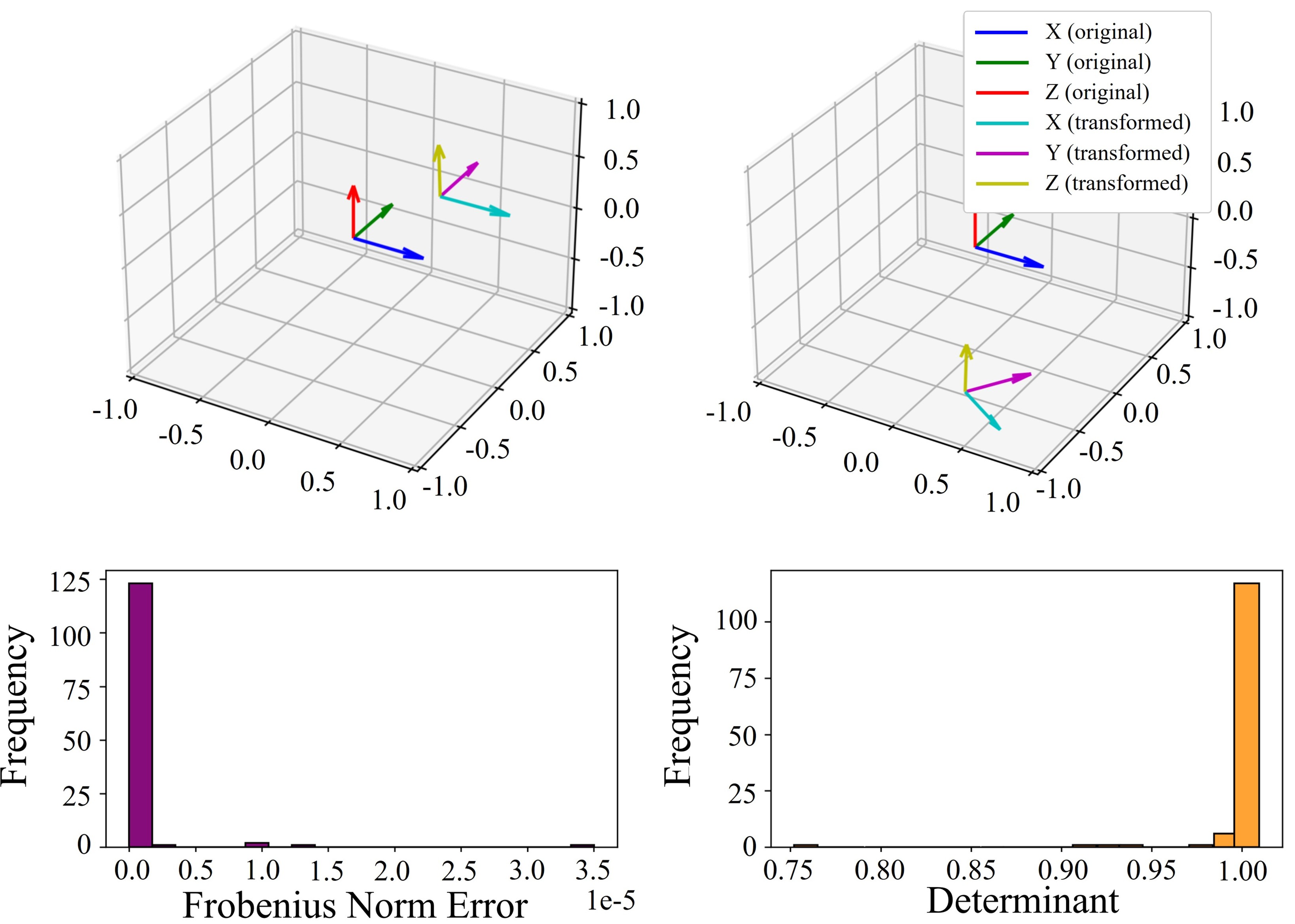}
    \caption{The visual and numerical results of the trained SEGGN. The first row illustrates the transformed coordinates based on the SEGGN output. The second row presents numerical evaluations, including the Frobenius norm error, calculated as $||\textbf{R}^\top\textbf{R} - \textbf{I}||_F$, and the determinant of \textbf{R}, both shown as histograms.}
    \label{fig:se_result}
\end{figure}

Furthermore, due to the structure of the SEGGN, which utilizes the bases of $\mathfrak{se}_2(3)$ followed by an exponential mapping process, it inevitably produces outputs within SE$_2$(3). The corresponding results are presented in Fig. \ref{fig:se_result}. The first row of Fig. \ref{fig:se_result} illustrates the transformation of the original coordinates based on the output values obtained from SEGGN. Visually, it can be observed that the transformed coordinates maintain orthonormality, ensuring the structural integrity of the transformation. A numerical analysis of these results is presented in the histograms below. The Frobenius norm error distribution shows that nearly all samples have errors close to zero, indicating minimal deviation from the expected transformation properties. Additionally, the determinant values are consistently close to one, confirming that the rotational component of the output matrix adheres well to the properties of the SO(3) group (\ref{eq:so3}). 
\begin{equation}
    \text{SO}(3)=\{\textbf{R}\in \mathbb{R}^{3 \times3} | \textbf{R}\textbf{R}^\top=\textbf{I}, \text{det}(\textbf{R})=1\}
    \label{eq:so3}
\end{equation}
These results demonstrate that SEGGN successfully generates outputs that satisfy the rotational and translational properties of the SE$_2$(3) group.

\begin{table}[t!]
\centering
\resizebox{0.8\columnwidth}{!}{%
\begin{tabular}{|c|c|c|c|}
\hline
\multirow{2}{*}{\shortstack{\textbf{Batch}\\\textbf{Size}}} & \multicolumn{2}{c|}{\textbf{Computation Time [ms]}} & \multirow{2}{*}{\shortstack{\textbf{Ratio}\\\textbf{(Add/Mul)}}} \\ \cline{2-3}
                            & \textbf{Addition}           & \textbf{Multiplication}        &                        \\ \hline
128                         & 0.0654            & 0.6140               & 0.1066                \\ \hline
256                         & 0.0703            & 0.6978               & 0.1008                \\ \hline
512                         & 0.0826            & 0.8612               & 0.0960                \\ \hline
1024                        & 0.1033            & 1.2170               & 0.0849                \\ \hline
\end{tabular}%
}
\caption{Computation time of addition and multiplication depending on batch size and their ratio.}
\label{table:add_mul_computation}
\end{table}

\subsection{Implementation Details}
\begin{figure*}
    \centering
    \includegraphics[width=\linewidth]{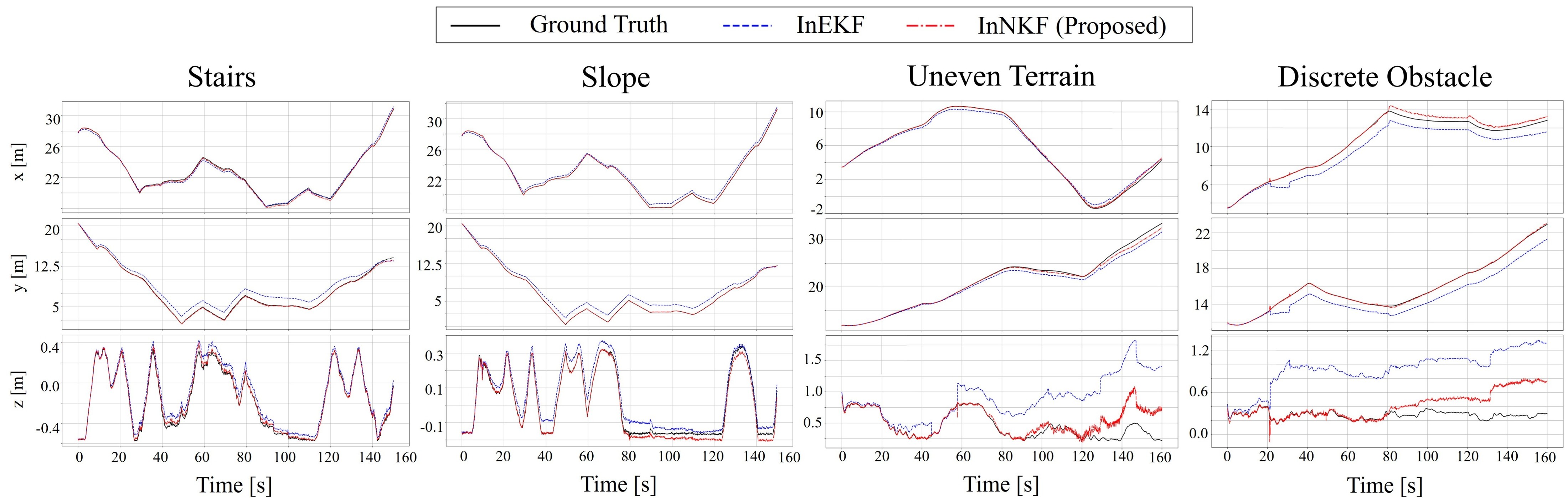}
    \caption{Position estimation results across four terrains (stairs, slope, uneven terrain, discrete obstacle) over a time interval of 0 $\sim$ 160 s. The solid black line represents the ground truth, the blue dashed line represents InEKF, and the red dashed line represents the InNKF (proposed method).}
    \label{fig:result_1}
\end{figure*}
To reduce the sim-to-real gap, sensor noise and IMU bias were introduced to reflect real-world conditions better. These noise factors were applied during both locomotion controller training and dataset collection. The sensor noise was assumed to follow a Gaussian distribution. Additionally, a covariance matrix was configured to address uncertainties in the measurement model. Since InEKF measurements primarily depend on kinematic information, the covariance matrix was constructed based on the standard deviation of the noise applied to joint position and velocity.
\subsection{Experimental Results}

\begin{figure}
    \centering
    \includegraphics[width=\linewidth]{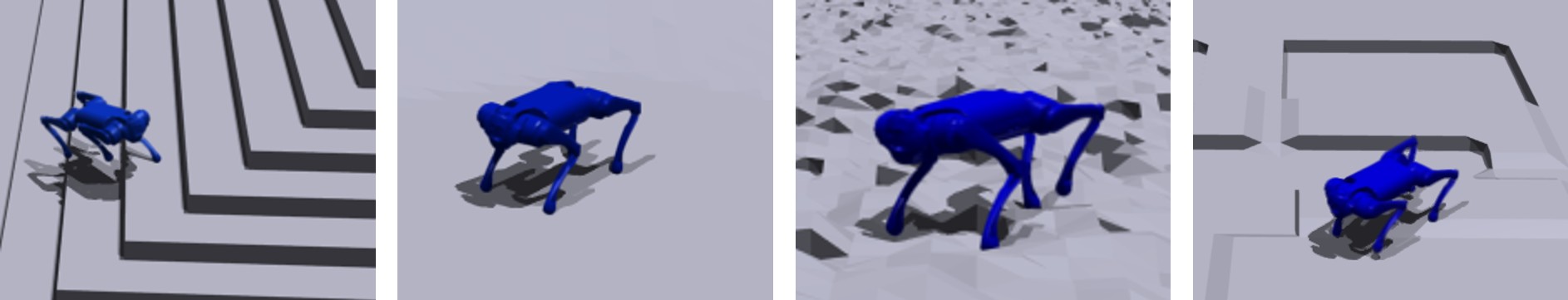}
    \caption{The four terrains (from left: stairs, slope, uneven terrain, and discrete obstacle) in the Isaac Gym are used for collecting training and test datasets.}
    \label{fig:isaacgym_screenshot}
\end{figure}
Three baselines were compared to analyze the experimental results. The first was the model-based (MB) estimator, represented by the Invariant Extended Kalman Filter (InEKF). The second was the learning-based (LB) estimator, represented by a Neural Network (NN), and the third was the hybrid estimator, represented by KalmanNet. 

\begin{figure}[t!]
    \centering
    \includegraphics[width=\linewidth]{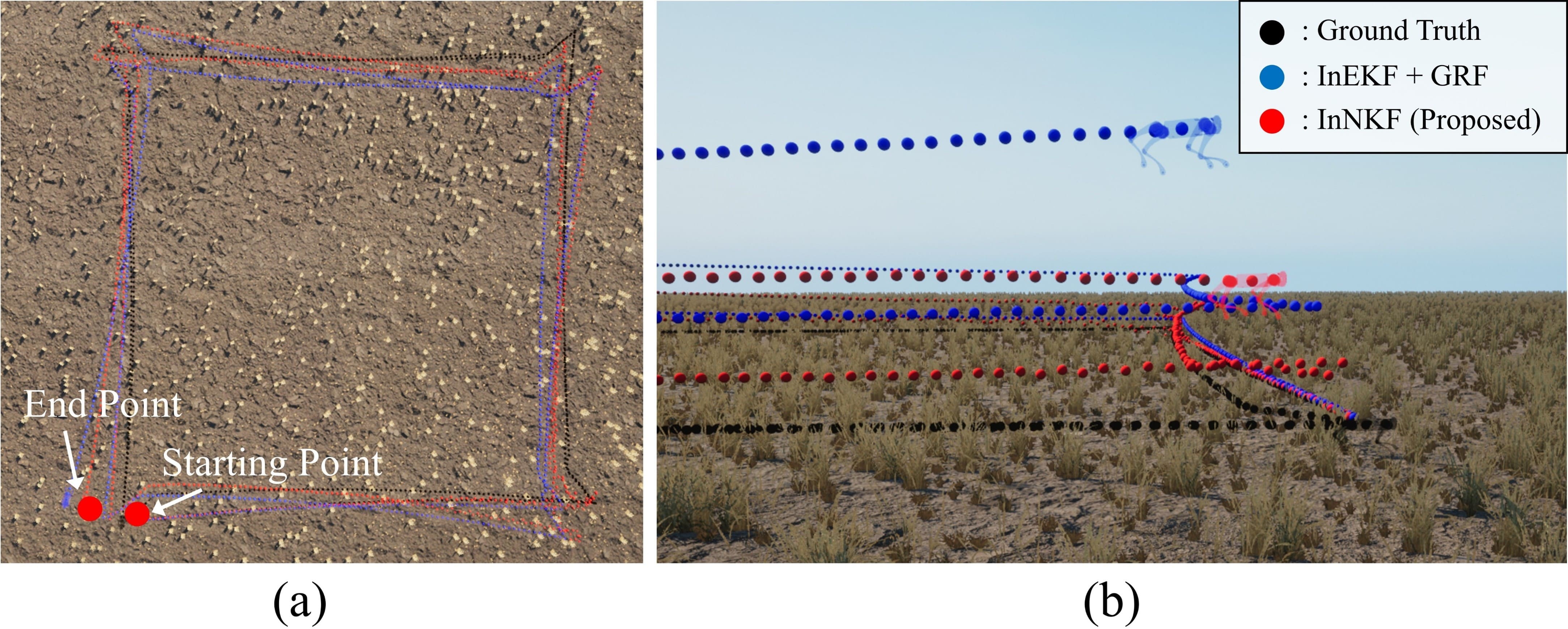}
    \caption{State estimation results in Raisim for two laps around a 20 m $\times$ 20 m square path on field terrain. The black line represents the ground truth, the blue line represents InEKF, and the red line represents InNKF (proposed). (a): Bird's-eye view, (b): Side view.}
    \label{fig:raisim_screenshot}
\end{figure}
\begin{figure}[t!]
    \centering
    \includegraphics[width=\linewidth]{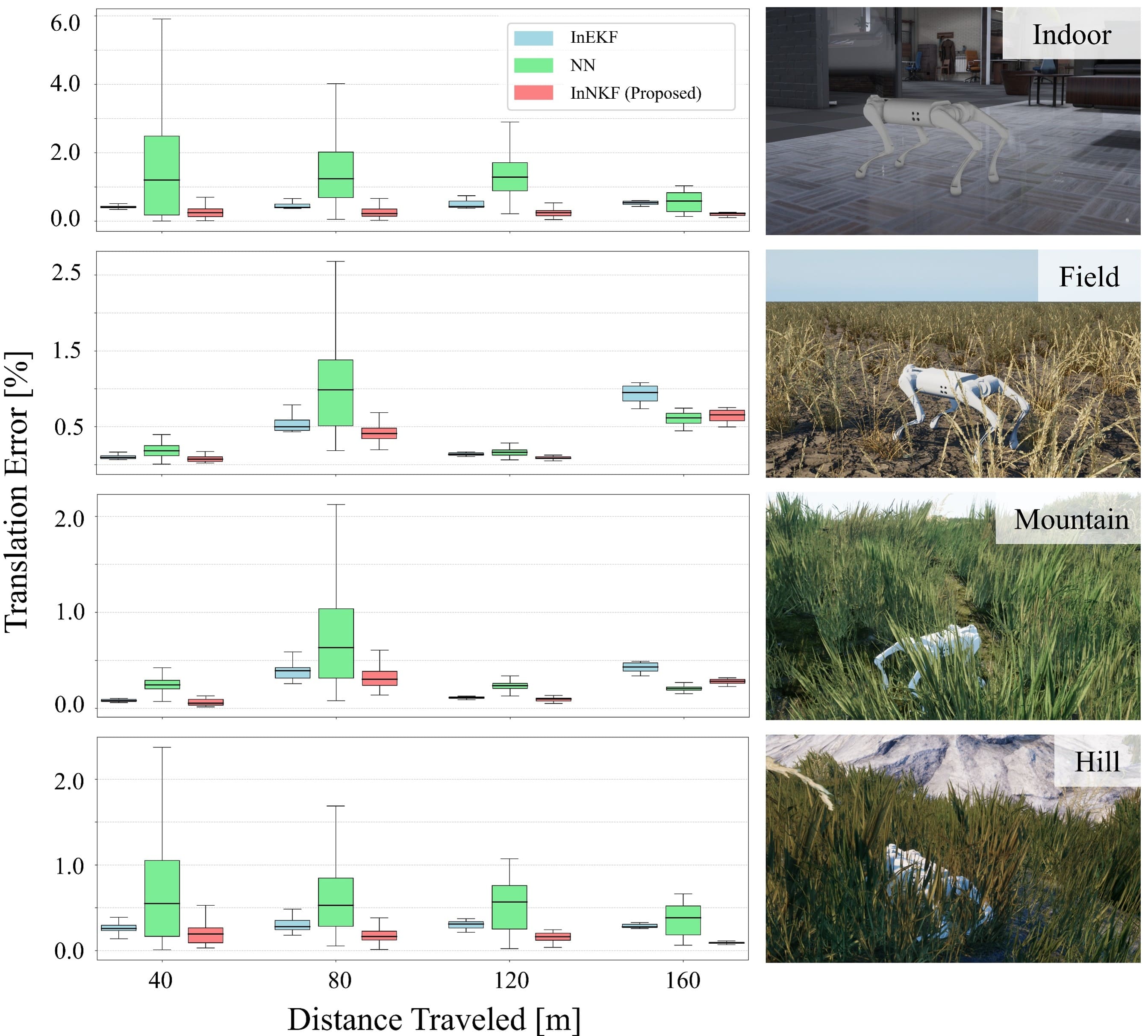}
    \caption{Relative Errors (RE) in Raisim across different environments (indoor, field, mountain, hill). In each box, blue represents InEKF, green represents NN, and red represents InNKF (proposed).}
    \label{fig:result_2}
\end{figure}
\begin{figure}[t!]
    \centering
    \includegraphics[width=0.8\linewidth]{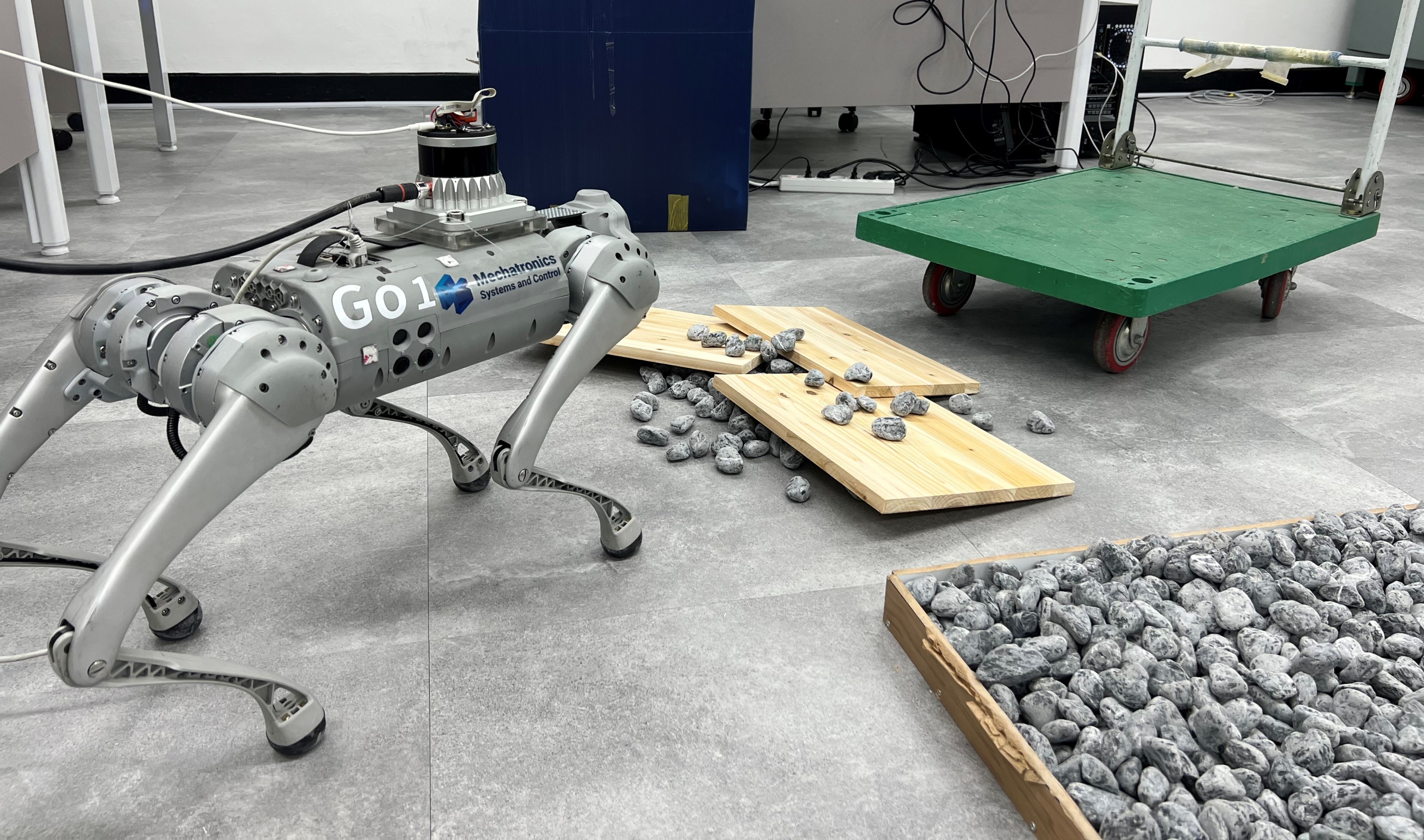}
    \caption{Real-world experimental setup with gravel, wooden planks, and a cart to create unstructured terrain.}
    \label{fig:real_world}
\end{figure}

The training and test datasets were collected in the Isaac Gym simulator \cite{makoviychuk2021isaac}, using terrains shown in Fig. \ref{fig:isaacgym_screenshot}. The quadruped robot, Unitree Go1, was the model used, controlled by a blind locomotion controller trained via reinforcement learning \cite{rudin2022learning}. To evaluate whether the state estimator overfits the training conditions, additional experiments were conducted in Raisim \cite{hwangbo2018per}. The results from Isaac Gym are presented in Fig. \ref{fig:result_1} and Table \ref{table:result_1}, while the results from Raisim are shown in Fig. \ref{fig:raisim_screenshot} and Fig. \ref{fig:result_2}. Furthermore, to assess real-world applicability, the proposed method was evaluated in environments depicted in Fig. \ref{fig:real_world}, with the corresponding results summarized in Table \ref{table:result_3}.

In Fig. \ref{fig:result_1}, the proposed method, InNKF, demonstrates a significant reduction in y-axis position error compared to InEKF. In particular, for the discrete obstacle terrain, the z-axis position error of InEKF is 0.3472m at 160s, whereas InNKF maintains an error of 0.1420m, confirming its improved performance. Table \ref{table:result_1} further evaluates the performance using Absolute Trajectory Error (ATE), including NN-based state estimation and KalmanNet which is the hybrid state estimation as the baselines. The results indicate that InNKF consistently reduces rotation ATE across all terrains. Specifically, velocity ATE is reduced by 40.1\% in the slope terrain, while position ATE is reduced by 87.0\% in the discrete obstacle terrain. These results demonstrate that NN-based state estimation alone does not provide reliable performance, whereas the proposed InNKF achieves superior accuracy in most cases. Also, the hybrid method KalmanNet showed good performance in rotation and position on the stairs terrain, but in most cases, it did not outperform InNKF.
\begin{table}[t!]
\centering
\resizebox{\columnwidth}{!}{%
\begin{tabular}{|c|c|c|c|c|c|}
        \hline
        \multirow{3}{*}{\textbf{Terrain}} & \multirow{3}{*}{\textbf{Eval. metric}} & \multicolumn{4}{c|}{\textbf{Algorithm}} \\
        \cline{3-6}
         &  & \textbf{InEKF} & \textbf{NN}  & \textbf{KalmanNet} & \textbf{InNKF} \\
         &  & \textbf{(MB-only)} & \textbf{(LB-only)} & \textbf{(Hybrid)} & \textbf{(Proposed)} \\
        \hline
        \multirow{3}{*}{Stairs}  
        & \textbf{ATE}\textsubscript{R} [rad] & 0.1247  & 0.4695 & \textbf{0.0809} & 0.0986  \\
        & \textbf{ATE}\textsubscript{v} [m/s]  & \textbf{0.1441}  & 0.2712 & 0.2579 & 0.2056  \\
        & \textbf{ATE}\textsubscript{p} [m]  & 0.6224 & 1.9160 & \textbf{0.3991} & 0.8697  \\
        \hline
        \multirow{3}{*}{Slope}  
        & \textbf{ATE}\textsubscript{R} [rad] & 0.0454  & 0.1212 & 0.0591 & \textbf{0.0231}  \\
        & \textbf{ATE}\textsubscript{v} [m/s]  & 0.0930  & 0.0648 & 0.0717 & \textbf{0.0373}  \\
        & \textbf{ATE}\textsubscript{p} [m]  & 1.0439  & 0.2363 & 0.5644 & \textbf{0.1455}  \\
        \hline
        \multirow{3}{*}{\shortstack{Uneven\\Terrain}}  
        & \textbf{ATE}\textsubscript{R} [rad] & 0.0388  & 0.3114 & 0.0502 & \textbf{0.0341}  \\
        & \textbf{ATE}\textsubscript{v} [m/s]  & \textbf{0.0930}  & 0.1744 & 0.3004 & 0.1186  \\
        & \textbf{ATE}\textsubscript{p} [m]  & 1.0401  & 1.4215 & 0.5834 & \textbf{0.4582}  \\
        \hline
        \multirow{3}{*}{\shortstack{Discrete\\Obstacle}}  
        & \textbf{ATE}\textsubscript{R} [rad] & 0.0500  & 0.1110 & 0.0732 & \textbf{0.0215}  \\
        & \textbf{ATE}\textsubscript{v} [m/s]  & 0.0958  & 0.0631 & \textbf{0.0263} & 0.0462  \\
        & \textbf{ATE}\textsubscript{p} [m]  & 1.0734  & 0.2036 & 0.6053 & \textbf{0.1389}  \\
        \hline
    \end{tabular}
}
\caption{Comparison of Absolute Trajectory Error (ATE) in state estimation among the three baselines and InNKF (proposed method) across different terrains.}
\label{table:result_1}
\end{table}

To validate its performance in a different dynamic simulator, position error was evaluated in Raisim, with additional data collected for fine-tuning. As shown in Fig. \ref{fig:raisim_screenshot}, when the robot completed two laps along a 20m $\times$ 20m square trajectory, the x and y-axis errors between the starting and ending points remained minimal. Additionally, InNKF demonstrated significantly improved accuracy along the z-axis compared to InEKF, staying closer to the ground truth. Fig. \ref{fig:result_2} compares the relative error (RE) across four environments—indoor, field, mountain, and hill—over a total walking distance of 160m. NN-based estimation exhibited high variance, reaching a maximum of 4.6298 in the indoor environment, whereas InNKF maintained a significantly lower maximum variance of 0.0466, highlighting the unreliability of NN-based estimation. As the distance traveled increased, InNKF consistently maintained a lower RE than InEKF, achieving up to 68.32\% performance improvement.

To verify the applicability of the proposed method in real-world environments, experiments were conducted in the setting shown in Fig. \ref{fig:real_world}. Since a motion capture system was unavailable, pseudo ground truth was obtained using LIO-SAM \cite{shan2020lio} with VectorNav VN-100 IMU and Ouster OS1-64-U LiDAR. Based on this, the state estimation performance was evaluated by comparing ATE and Relative Error (RE), and the results are summarized in Table \ref{table:result_3}. The proposed InNKF demonstrated improvements of up to 77\% in ATE, confirming its feasibility for real-world applications.
\begin{table}[t!]
\centering
\resizebox{\columnwidth}{!}{%
\begin{tabular}{|c|c|c|c|c|}
\hline
\multirow{3}{*}{\textbf{Eval. metric}} & \multicolumn{4}{c|}{\textbf{Algorithm}} \\
\cline{2-5}
& \textbf{InEKF} & \textbf{NN} & \textbf{KalmanNet} & \textbf{InNKF} \\
& \textbf{(MB-only)} & \textbf{(LB-only)} & \textbf{(Hybrid)} & \textbf{(Proposed)} \\
\hline
\textbf{ATE}\textsubscript{R} [rad] & 0.2564 & 0.1453 & 0.1627 & \textbf{0.0330} \\
\textbf{ATE}\textsubscript{p} [m] & 0.2042 & 0.1291 & 0.2166 & \textbf{0.0726} \\
\hline
\multirow{2}{*}{\textbf{RE}\textsubscript{R} [rad/m]} & 0.0210 & 0.0220 & 0.0211 & \textbf{0.0209} \\
& (0.0021) & (0.0114) & (0.0043) & \textbf{(0.0021)} \\
\multirow{2}{*}{\textbf{RE}\textsubscript{p} [\%]} & \textbf{1.0019} & 3.3852 & 1.2109 & 1.1690 \\
& \textbf{(0.1727)} & (2.0124) & (0.2729) & (0.4459) \\
\hline
\end{tabular}
}
\caption{ATE and 10 Seconds RE (Mean and Standard Deviation) of state estimation in the real world.} 
\label{table:result_3}
\end{table}
\section{CONCLUSIONS}
\label{section:conclusion}
This study proposed the Invariant Neural-Augmented Kalman Filter (InNKF), a novel state estimator that integrates a neural network-based compensation step into a model-based framework. Using a neural network as a nonlinear function approximator, InNKF corrects linearization errors at each time step, enhancing state estimation performance. However, as it relies on dataset-based learning, retraining is needed for different robot models, and the lack of a motion capture system in real-world experiments limited precise evaluation. SEGGN operates at 660 Hz, suggesting that its integration with InEKF (up to 2000 Hz) ensures real-time capability. Future work will refine the Neural Compensator structure, address key nonlinear factors, and extend its applicability to various legged robots.







\bibliographystyle{ieeetr}

\end{document}